\documentclass[letterpaper, 10 pt, conference]{ieeeconf} 
\IEEEoverridecommandlockouts                              
\usepackage{graphicx}
\usepackage{amsmath}
\usepackage{amssymb}
\usepackage{booktabs}
\usepackage{multicol}
\usepackage{multirow}
\usepackage{nicefrac}
\usepackage{siunitx}
\usepackage{numprint}
\usepackage[pagebackref,breaklinks,colorlinks]{hyperref}
\usepackage[flushleft]{threeparttable}
\usepackage{verbatim} 

\title{\LARGE \bf
Mainline Automatic Train Horn and Brake Performance Metric*
}

\author{Rustam Tagiew$^{1}$  % 
\thanks{*This work originated within the tasks of the Network of Experts of the German Federal Ministry for Digital and Transport. This is not an official statement, guideline or directive of the German Federal Railway Authority.}% 
\thanks{$^{1}$Rustam Tagiew is a scientific desk officer at German Centre for Rail Traffic Research (DZSF) at Federal Railway Authority, August-Bebel-Str. 10, 01219 Dresden %
        {\tt\small info@dzsf.bund.de}}%
}
\newcommand{\rem}[1]{}
\begin{document}

\maketitle
\thispagestyle{empty}
\pagestyle{empty}

\begin{abstract}
This paper argues for the introduction of a mainline rail-oriented performance metric for driver-replacing on-board perception systems. Perception at the head of a train is divided into several subfunctions. This article presents a preliminary submetric for the obstacle detection subfunction. To the best of the author's knowledge, no other such proposal for obstacle detection exists. A set of submetrics for the subfunctions should facilitate the comparison of perception systems among each other and guide the measurement of human driver performance. It should also be useful for a standardized prediction of the number of accidents for a given perception system in a given operational design domain. In particular, for the proposal of the obstacle detection submetric, the professional readership is invited to provide their feedback and quantitative information to the author. The analysis results of the feedback will be published separately later.     
\end{abstract}

\section{INTRODUCTION}

Driverless and unattended train operations show multiple advantages \cite{ATOadvantages}. These advantages are increases in capacity, reliability, service flexibility and energy efficiency as well as alleviating the shortage of train drivers. So far, these advantages can only be enjoyed in the case of fully automated metros in regular operation. Driverless train operation for mainline trains is still an unsolved challenge. The crucial difference is that in mainline railways, the tracks are open to disruptive exogenous influences, and the use of track-side measures such as fencing and cab signaling is not economically justifiable.

The challenge to be solved for mainline railway automatization is related to the fully automated road traffic and can benefit from technology transfer. It requires a development of a vehicle-side AI system performing a multi-sensory perception. However, a literature review has shown an order of magnitude lower research activity for rail traffic in comparison to road traffic \cite{railwayvisonreview}. It also showed insufficient progress -- the current Technology Readiness Level (TRL) for rail traffic is 5 and remains unsurpassed for the last two decades. A key finding was the absence of a widely accepted performance metric that could link rail safety requirements with AI developers' community. 

This article attempts to solve the issue with the absent performance metric. Such a performance metric would, on the one hand, provide developers with clear application-oriented goals, make their results comparable and, on the other hand, make progress measurable for outsiders. The article proposes a preliminary performance submetric for the major subfunction of obstacle detection. Based on this proposal, a discussion can be initiated and first perception system performance results can be compared. The readers of this paper are urged to actively submit either the performance data of their systems or their suggestions for improvement to the author. Further, this paper also aims to bring the domain inexperienced developers closer to the condition of railway to increase the research facilitating effect. 

\section{DIVISION INTO SUBFUNCTIONS}
\label{overview}
\begin{figure}
  \centering
  \includegraphics[width=1\linewidth]{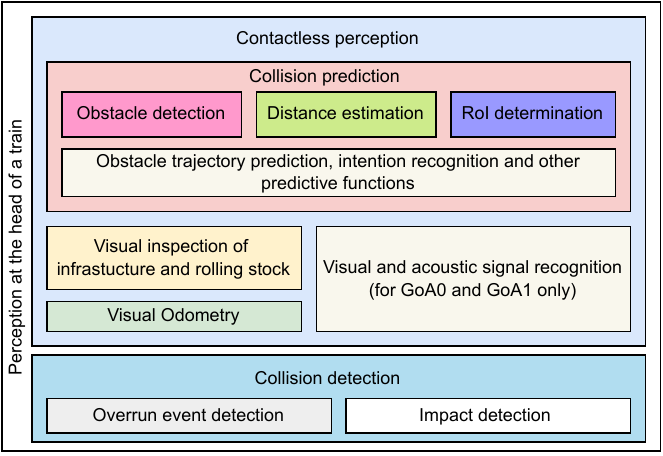}  
  \caption{Rough categorization of system's functions for driverless mainline rail traffic, which in comparison to driverless metros, require additional technological effort \cite{railwayvisonreview}.}
  \label{trainside}
\end{figure}
AI systems replacing human staff on trains have to perform multiple functions. Considering current state-of-art, these functions will not represent one-to-one the full range of human abilities, they only cover the most relevant tasks on at least same performance level. In comparison to state-of-art driverless metros, the system replacing the train driver on mainline railways requires extra development. The functions can be divided into mainly two subfunctions, the perception of objects with and without physical contact (fig.\ref{trainside}). Fig.\ref{trainside} does not include many subfunctions like e.g. surveillance of door operation, emergency detection, crime detection and so on.

Perception by contact with objects is referred to here as collision detection and replaces the train driver's acoustic and haptic sensation\rem{ sometimes referred to as a ``popometer''}. Already in EN 62267, a standard for driverless metros, it is mentioned that a collision has to be detected at the latest at the contact with an obstacle. In the special case of shunting, controlled collisions such as running into a drag shoe or coupling of cars are part of normal operation. In all other cases, collisions with objects are unwanted, dangerous accidents that cannot always be avoided and must always be detected. Two types of collisions can be identified so far, impact and overrun events. The detection of impact events is referred to here as impact detection. For mainline railways, little research on impact detection and only one seminal research on overrun event detection systems \cite{overrunevent} are known. Collision detection is therefore at a very early stage of development.  

Detection of obstacles without physical contact replaces the human sight from the cab. It includes multiple tasks having driveway surveillance for collision prediction as the most challenging of them \cite{atosensorikreport}. It is assumed that collision prediction is always prone to errors, false negatives and false positives, and therefore cannot make collision detection obsolete.

Visual inspection of infrastructure and rolling stock is more important for mainline railways than for metros due to greater exogenous influences and bigger operational areas, and is not only important for predictive maintenance. There are also cases such as sun kinks, catenary damage, broken signals, malfunctioning railroad crossing gates and slipping loads during train meets that require emergency braking and are therefore part of the driving function. Visual odometry complements rotary encoders, Inertial Measurement Units (IMU) and sensors for Global Navigation Satellite System (GNSS). 

The railway signals have to be recognized from the vehicle. There are multiple groups of signals, which can be visual or acoustic. In case of shunting for the lowest Grade of Automation (GoA) 0, signals are e.g. fouling point indicators at the railway switches. Although the detection of signals is ensured by automatic train stop in case of GoA1, they still have to be recognized from the vehicle. The challenge of signal detection also includes detection of tracks and their assignment to the signals \cite{petrovic2022integration, staino2022real}. From GoA2 on, signals do not need to be detected and are transmitted by cab signalling when used with ETCS. The GoA2 can also be conceptually achieved if an automatic visual detection of signals assists the driver \cite{ei1}.

Prediction of collision with obstacles requires algorithms for obstacle detection, distance estimation, Region of Interest (RoI) determination, obstacle trajectory prediction, human intention recognition and other predictive functions. Depending on the choice of operational design domain (ODD), some of the functions, such as human intention recognition, may be unnecessary. Obstacle detection can be further subdivided into object detection, obstacle classification and spatial angle determination. There are internal obstacles such as railway vehicles and buffer stops. The external obstacles can be pedestrians, road cars, big animals, trees, rocks, wrongly placed drag shoes, floods, fires and similar. Obstacles do not only appear on the ground, they might also hang on the catenary \cite{fahrd1,fahrd2,fahrd3,fahrd4,fahrd5} or levitate in the air.

Distance estimation is important for shunting and also for detecting obstacles from long distances in curves, where a relatively small distance error determines whether or not an object intersects with the structure gauge \cite{gebauer2012autonomously}. Spatial angle determination together with distance estimation is referred to as obstacle localization. For the RoI determination, a 3D tubular space formed by the predicted train's driveway and the structure gauge should be determined in the scene. Train's driveway is also known as train's path \cite{ristic2021review}. The structure gauge is supplemented with a speed-dependent hazard area for pedestrians, which arises due to aerodynamics around a moving train \cite{guv}. In the rare case that the states of the switches are not otherwise available to the perception system, they must be extracted from the visual input for the train's path prediction. 
\section{SAFETY ARGUMENTATION}
\begin{figure}
  \centering
  \includegraphics[width=1\linewidth]{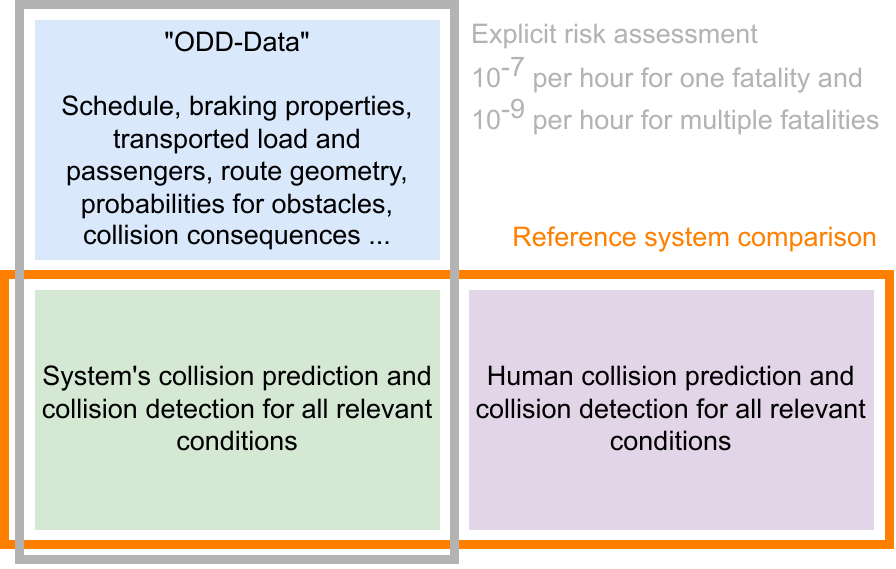}  
  \caption{Data required for two currently available approaches of safety argumentation for European mainline railway systems \cite{riskanal}. The grey frame denotes the explicit risk assessment with resulting hourly fatality rates and the maximal values of harmonized design goals. The orange frame denotes the comparison with the a human train driver as reference system.}
  \label{csmra}
\end{figure}
All subfunctions, described in Sec.\ref{overview}, require performance indicating submetrics for all relevant stakeholders, especially the developers and the regulators. Safety relevant functions for European mainline railways are approved according to the Common Safety Method for Risk evaluation and Assessment (CSM-RA). Hereby, performance metrics are needed, which allow prove of compliance with standards, comparison with human performance or calculation of resulting hourly fatality rates. Since there are still no standards for this, only two remaining approaches of safety argumentation are available (fig.\ref{csmra}). These are the reference system comparison and explicit risk assessment according to harmonised design goals. 

As depicted in fig.\ref{csmra} for collision risks, both approaches need performance data of system's collision prediction and collision detection for all relevant conditions. Explicit risk assessment requires additional data to calculate, whether the probability for an accident with a single fatality is lower than $10^{-7}$ and for an accident with more than one fatality is lower than $10^{-9}$. This additional data includes schedule, braking properties, route geometry, probabilities for obstacles, collision consequences, acoustic properties for warning horn, transported load and passengers. This data describes the ODD of a train and is called here ''ODD-Data''. Instead of ODD-Data, the reference system comparison needs performance data of human collision prediction and collision detection for all relevant conditions.
 
\begin{figure*}
  \centering
  \includegraphics[width=1\linewidth]{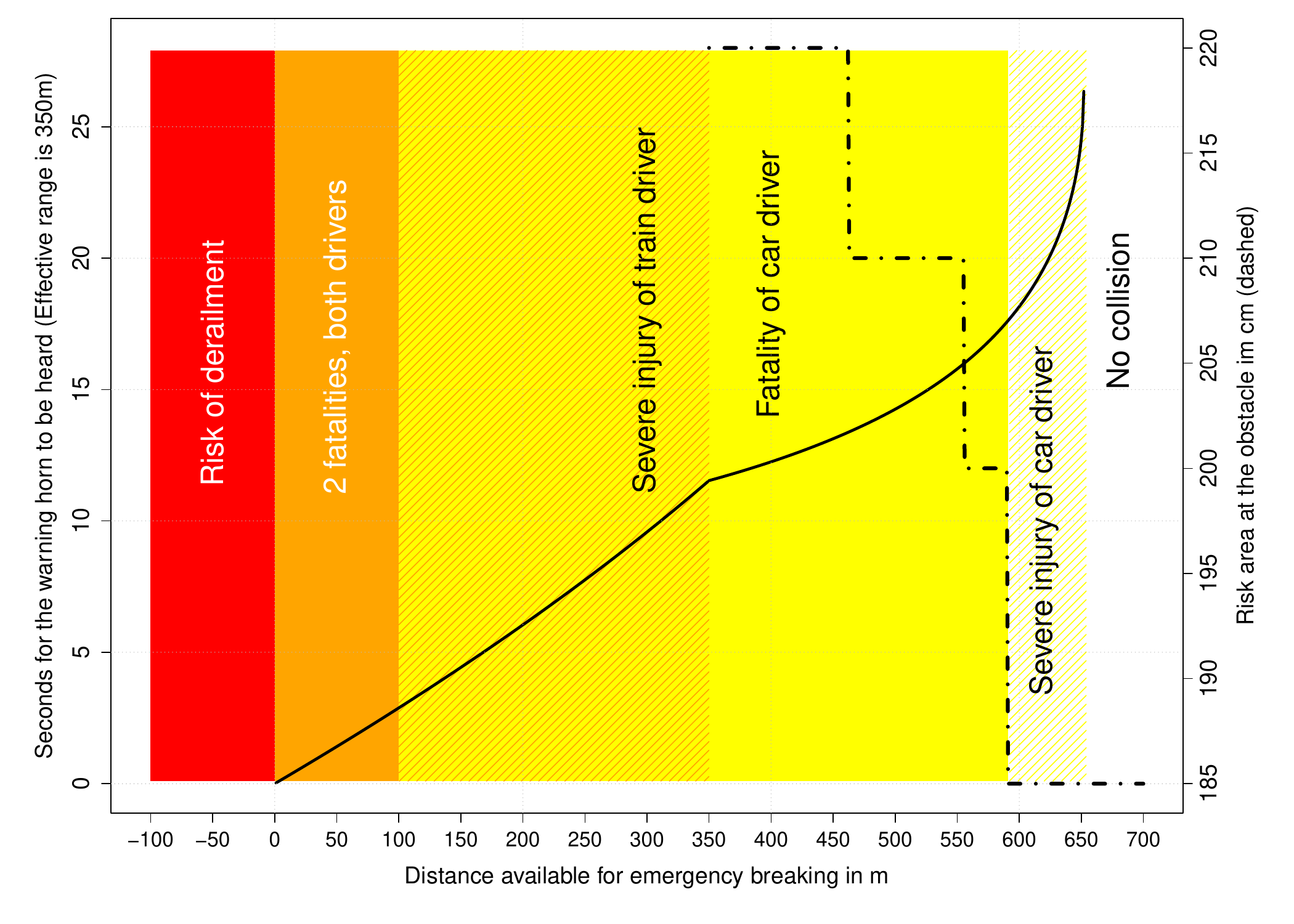}  
  \[ t = \frac{1}{a}\left(\left( \begin{cases}
        v                  &  \SI{0}{\meter}\leq d<\SI{350}{\meter}  \\
        \sqrt{v^2-2a(d-\SI{350}{\meter})} & \SI{350}{\meter}\leq d\leq\SI{652}{\meter}
    \end{cases}\right) - \sqrt{v^2-2ad}\right) \text{ ; $t$-time, $a$-decceleration, $v$-speed, $d$-distance} \]
  \caption{Estimated consequences for a frontal collision of a train going at $\SI{130}{\nicefrac{\km}{\hour}}$ with a stationary passenger car depending on braking distance. The braking deceleration is set to be $\SI{1}{\nicefrac{\meter}{\second^2}}$. The driver can hear the warning horn at a distance of $\SI{350}{\meter}$ or less and may be able to escape. Negative distances mean the onward movement of an unbraked collided train. Warning horn and emergency braking start simultaneously. The solid kinked curve shows the number of seconds between hearing the warning horn by the car driver and the collision. The formula for this curve is added below the graph and provides an explanation for the kink. The dashed zigzag line depicts the size of the risk area at the collision site. For the sake of simplicity, it is assumed in this that the derailment risk in this example is only present in collisions at speeds of $\SI{130}{\nicefrac{\km}{\hour}}$ and above.}
  \label{risk}
\end{figure*}

\section{REQUIREMENTS FOR OBSTACLE DETECTION}
To justify performance metrics, a detailed description of obstacle detection in the railway domain is given with a focus on safety. Commonly used performance metrics do not correlate well with the safety argumentation. In particular, the performance metric Intersection over Union (IoU), which is oriented to the 2D space of camera images, could mislead the development of a perception system. Even in 3D space, IoU still requires a safety-oriented weighting of the spatial direction of the mismatch between prediction and ground truth. Mean Average Precision (mAP) based on IoU provides a value only for single shot prediction, not for a sequence of images of a train approaching an obstacle.     

As according to the statistics of Eurostat for 2021 in EU \cite{eurostat}, $\SI{64.5}{\text{\%}}$ of fatalities result from accidents to persons by rolling stock in motion, $\SI{34.3}{\text{\%}}$ from level crossing accidents including pedestrians and only $\SI{1.2}{\text{\%}}$ from railway vehicle collisions and other accidents. The portion of pedestrians in level crossing accidents can be assumed to be $\SI{14.6}{\text{\%}}$ based on German statistics by Deutsche Bahn \cite{dbstats} for 2018. Therefore, the most probable fatal accident scenario is collision with a person at roughly $\SI{70}{\text{\%}}$. Second most probable scenario is collision with a passenger car at roughly $\SI{24}{\text{\%}}$. 

Both most common scenarios on railways, pedestrian- and car-collisions, are also most common on roads \cite{caraccident}. In contrast to commonly known road vehicles, emergency braking and warning horn are the only available reactions on railway. The braking distance for railway vehicles is approximately $5$ times longer than for road vehicles. The ca. $\SI{15}{\text{dB(A)}}$ louder warning horn can and should be heard from larger distances \cite{autohupe,whistledistance}. Both has consequences on the minimal acceptable performance of vision-based collision prediction and makes Long-Range Object Detection (LROD) necessary. Due to curvatures, weather and light conditions, LROD is not always possible. While for road vehicles, collision prediction enables its prevention, it is rather a matter of damage limitation and reverence in the domain of railway vehicles. 

Collision with a person causes a fatality for all ego vehicle speeds in case of railways as according to DIN VDE V 0831-103. However, out of a total of $695$ accidental fatalities and serious injuries in 2021 in EU caused by rolling stock in motion, $\SI{36,5}{\text{\%}}$ were seriously injured, i.e. survived \cite{eurostat}. When a deadly collision with a person cannot be prevented, the braking must be applied to preserve the dignity of the dead, to facilitate investigation by authorities and prevent exposure to casual bystanders. This is also important for the more than $\numprint{2000}$ rail suicides in EU each year, which are not counted as accidents. Warn horn and braking is never too late and has to be done as soon as possible in this scenario.

\begin{table}%[bh]
  \caption{Human detection of objects on railways in m.}
  \label{distances}
  \begin{center}
  \begin{tabular}{|l||r|}
    \hline
    Object & Median distance \\
           & of detection \\
    \hline
    $\SI{0.4}{\meter^2}$ and $\SI{2}{\meter^2}$, $\SI{30}{\text{\%}}$ contrast & $>750$ \\
    $\SI{2}{\meter^2}$, $\SI{8}{\text{\%}}$ contrast & $500$  \\
    $\SI{0.4}{\meter^2}$, $\SI{8}{\text{\%}}$ contrast & $240$ \\
    $\SI{2}{\meter^2}$, $\SI{30}{\text{\%}}$ contrast, at night  & $180$ \\
    $\SI{0.4}{\meter^2}$, $\SI{30}{\text{\%}}$ contrast, at night  & $60$ \\
    $\SI{0.4}{\meter^2}$ and $\SI{2}{\meter^2}$, $\SI{8}{\text{\%}}$ contrast, at night & $<60$ \\
    \cite{polz} & \\
    \hline
    $40\times40\times\SI{40}{\text{cm}}$ & $250$ \\
    $20\times20\times\SI{20}{\text{cm}}$ & $175$ \\
    $10\times10\times\SI{10}{\text{cm}}$ & $50$ \\
    $5\times5\times\SI{5}{\text{cm}}$ & $<25$  \\
    fluorescent objects at night, $\SI{60}{\nicefrac{\km}{\hour}}$ \cite{itoh} & \\
    \hline
    person in safety jacket & $400$ \\
    passenger car & $300$ \\
    person & $240$ \\
    passenger car at night & $<60$ \\
    person in safety jacket at night & $<60$ \\
    person at night & $<60$ \\
    \cite{mockel2003multi} & \\
    \hline
    trees, $50$-$\SI{70}{\nicefrac{\km}{\hour}}$ & $60$ \\
    fallen rocks, $20$-$\SI{120}{\nicefrac{\km}{\hour}}$ & $30$ \\
    accident statistics \cite{nakasone2017frontal} & \\ 
    \hline
  \end{tabular}
  \end{center}
\end{table}
Collision with a passenger car is more intricate scenario than with a person regarding the consequences of different ego vehicle speeds. Fig.\ref{risk} shows the roughly estimated consequences for the collision of a train travelling at $\SI{130}{\nicefrac{\km}{\hour}}$ with a stranded passenger car. For simplicity, a uniform emergency braking deceleration of $\SI{1}{\nicefrac{\meter}{\second^2}}$ without delay is assumed. More realistic modeling would require consideration of additional modifiers such as co-functioning of different types of brakes, sanding to improve adhesion, and surge behavior of the liquid load. In the best case, if the car is recognised at more than $\SI{652}{\meter}$, the emergency braking can prevent the collision. In the worst case, if the car is not recognised before the collision, the impact detection system should recognise the crash and break to reduce the risk of a potential derailment of the train. The LROD can not always achieve the best case due to obstruction of view in curves, through hilltop, weather conditions, insufficient illumination, as well as due to sudden intrusion of a moving obstacle. 

However, earlier braking between the best and worst case reduces harm, which can be shown in our example in fig.\ref{risk}. According to the risk model by ENOTRAC \cite{moser2017analyse}, the damage of obstacles to a train grows with their mass and the speed of the train. According to DIN VDE V 0831-103, a crashing train with a speed higher than $\SI{40}{\nicefrac{\km}{\hour}}$ will cause fatality of the car driver. If the car driver can escape the car after hearing the warning horn, early braking gives more time for the resort depicted as solid curve. The assumption for the maximal distance of $\SI{350}{\meter}$ at which the warning horn can be heard by the car driver is derived from the German regulation for the maximal distance between a railroad crossing and a whistle board \cite{signalbuch}. A lower speed at the obstacle as a consequence of early braking reduces the risk area created by air stream around the vehicle depicted as dashed zigzag line as according to the speed thresholds in the regulation of German Statutory Accident Insurance \cite{guv}.

The distribution of distances, at which human drivers detect objects on railway, has an irregular bell shape \cite{riskanal}. Tab.\ref{distances} shows median distances for human performance at detecting objects on the tracks from all known sources. According to these measurements, a human driver can prevent collision with the car only if the car is of contrasting paint and is presented at daylight. At night without illumination, rainy weather and a decent car paint, the consequences will be much more severe. The shapes of the obstacle detection distances distribution are much steeper for computer vision systems than for humans \cite{mockel2003multi,nakasone2017frontal}. One source reports distances \cite{zhangyu2021camera}, at which first more or less erroneously placed boxes appear for target objects.\\
\indent False-negative and false-positive obstacle detection might occur due to reproducible or irreproducible failures in sensors and algorithms. The failures can be assigned to certain functions in certain cases. For instance, objects like stones and trees from the perceived space outside of RoI can be detected as obstacles due to wrong localization of them or to wrong localization of RoI. The space perceived by sensors is often larger than the 3D RoI, even in the presence of view obstructions. Another example is small animals that are recognized as obstacles because they are misclassified as humans or vice versa.\\
\indent Computationally, moreover, false-positive visual detection, i.e., false alarms, must occur much less frequently than false-negatives, since the case of absent obstacle is overwhelmingly predominant and obstacles are extremely rare. Additionally, mainline railway vehicles' emergency breaks can not be interrupted until full stop in many cases, create jams, damage to the vehicles and constitute therefore a significant cost factor, which has to be considered in the performance metrics. Since false-positive detection can not be outruled, collision prediction will be most probably complemented by impact detection to refute false visual detection \cite{riskanal}. 
\begin{figure*}[!ht]
  \centering
  \includegraphics[width=1\linewidth]{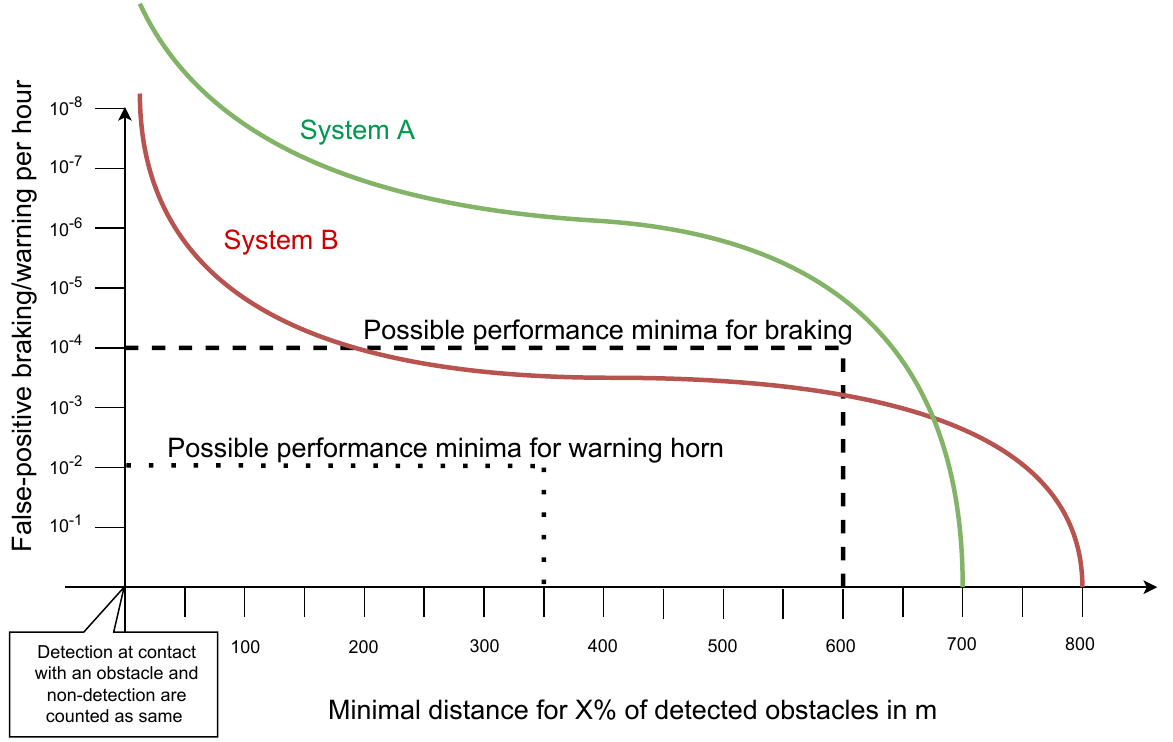}  
  \caption{Performance submetric for obstacle detection with results of two hypothetical systems A and B. $X$ can be replaced by a positive number up to $100$. A detection on contact with an obstacle and a non-detection are counted as the same.}\label{metrics}
\end{figure*}
\section{PROPOSED PERFORMANCE SUBMETRIC}
Fig.\ref{metrics} shows the proposed obstacle detection metric. This metric is designed for moving train. The abscissa shows the distances, at which $X\text{\%}$ of appearing obstacles are detected while approaching them. $(100-X)\text{\%}$ are detected at closer distances. $X=50$ denotes a median distance for obstacle detection. The ordinate shows hourly rates of false-positive detections, which will cause unneeded warning horn and jam-creating emergency braking. The values on the ordinate are negative logarithms of the hourly rate, the lower the better. The performance values of a system on these two axes are interlinked and can be adjusted by changing detection thresholds and tweaking internal parameters of a system. Like with Precision-Recall (PR) and Receiver Operating Characteristic (ROC) curves, increasing performance on the one axis will most probably reduce performance on the other axis.

The results according to this metric depend on the number and type of obstacles, the speed of the ego vehicle, the frame rate of the sensors, the track geometry, the time of day, the weather conditions, and other properties of a data set used to validate a system. The characteristics of the validation dataset will most likely depend on a chosen ODD. The shape of such performance curves is speculative and is shown in Fig.\ref{metrics} for hypothetical systems A and B. 

Both systems A and B have maximum ranges due to their sensor resolutions. Setting the internal thresholds of one system to the extreme of permanent positive detection will give the maximum range on the abscissa and $10^0$ on the ordinate. The opposite extreme, where the system is in permanent negative detection, will result in $0$ on the abscissa and $10^{-\infty}$ on the ordinate. The shapes of the curves in between for the hypothetical systems are drawn based on intuition. From the shape of the curve for the system A, it can be inferred that $(100-X)$\% of car-collision scenarios will result in one or more fatalities with this system adjusted to $<10^{-4}$ false-positives (fig.\ref{risk}). The system B has to be adjusted to $<10^{-3}$ false-positives, 10x more inappropriate stops, to achieve the same level of safety.

Based on a certain ODD, there will be certain performance minima for each of the axis. If the functions of emergency braking and warning horn are separated, the performance minima for both functions can be different. In project KOMPAS, $10^{-4}$ or less false-positive emergency braking per hour is suggested as the minimal acceptable performance \cite{polz}. Since false-positive warning horn does not create jams on the railways, the minimal requirements can be much less rigorous. However, extensive false-positive warning horn will probably not be welcomed by residents living close to the railway. For orientation, this paper proposes a rate of $10^{-2}$ cases per hour as depicted in fig.\ref{metrics}. 

The issue with the minima for distances depends stronger on ODD. Certain ego vehicle speeds, driveway geometries, weather and illumination conditions either prohibit or do not demand LROD for safety argumentation. For instance in case of car-collision scenario, warning horn is assumed to be effective a most $\SI{350}{\meter}$ only. Low ego vehicle speeds or better brakes result in lower distance requirements for obstacle detection. If a typical curved route does not allow sensors to penetrate further forward than $\SI{600}{\meter}$, a system will not be required to have a higher range. Both minima are depicted in fig.\ref{metrics}.      

In the pedestrian-collision scenario, the emergency braking function demands a system to overcome simultaneously higher minima on both axes than the warning horn function. In such case, system A is better than system B for both functions. For the pedestrian-collision scenario, effective distance for warning horn can be significantly longer than braking way \cite{whistledistance,Toward22} and that can make system B more appropriate for warning horn subfunction, while system A is more appropriate for emergency braking subfunction. 

Once the performance minima are met, the order of preference for both performance values becomes important in the choice of system and system parameter configuration. This could lead to answers to questions such as how much resident annoying extra warning horn is justified to save the life of one unlawful trespasser or one wild animal. 
\section{CONCLUSION AND CALL FOR DATA}
A very important idea of this work is the inaptitude of the concept of a binary false negative rate for obstacle detection for mainline railways. The non-detection of obstacles is gradual and not binary. The question is not ``What percentage of the obstacle is detected?''. The question is ``At what distance will $X\text{\%}$ of the obstacles be detected at the latest?''. The other important idea is that driving or minimizing the amount of false-positive stops is the primary goal and computationally more challenging, while safety or maximizing the timeliness of obstacle detection is the secondary goal. 

This paper is intended to elicit feedback from the research community. It contains a proposal for a submetric for an autonomous train perception system and a rationale for its design. The amount of feedback will be maximized by wide dissemination. The data expected here are lists of measurements that fit within the proposed submetric in fig.\ref{metrics} and $4$-tuples of the performance minima for braking and warning. An element in the list of measurement contains the name of the system, the $X$, rate of false-positives per hour and the minimal distance for $X$\% detections. Textual feedback is also welcome, especially as reasoning for the suggested performance minima. Also, human performance measurements as benchmark are welcome. The anonymized data from the feedback will be analyzed and published in a separate paper, for which this paper serves as a draft.

{\small
\bibliographystyle{ieee_fullname}
\bibliography{mawb}
}
\end{document}